%% file: main.tex
\documentclass[lettersize,journal]{IEEEtran}

\IEEEoverridecommandlockouts                      
\title{\LARGE \bf
UniFucGrasp: Human-Hand-Inspired Unified Functional Grasp Annotation Strategy and Dataset for Diverse Dexterous Hands}

\author{Haoran Lin$^{1,2,*}$, Wenrui Chen$^{1,2,}$\textsuperscript{\textdagger}, Xianchi Chen$^{1,*}$, Fan Yang$^{1,2}$, Qiang Diao$^{1}$,\\Wenxin Xie$^{3}$, Sijie Wu$^{3}$, Kailun Yang$^{1,2}$, Maojun Li$^{3}$, and Yaonan Wang$^{1,2}$%
\thanks{This work was partially supported by the National Key R\&D Program of China under Grant 2022YFB4701400/2022YFB4701404, the National Natural Science Foundation of China under Grant 62273137, 62473139, No. U21A20518, and No. U23A20341, the Hunan Provincial Research and Development Project under Grant 2025QK3019, the Hunan Science Fund for Distinguished Young Scholars under Grant 2024JJ2027, the Open Research Project of the State Key Laboratory of Industrial Control Technology, China (Grant No. ICT2025B20), and the State Key Laboratory of Autonomous Intelligent Unmanned Systems (the opening project number ZZKF2025-2-10).
}
\thanks{$^{*}$Equal contribution. \textsuperscript{\textdagger}Corresponding author.}%
\thanks{$^{1}$The authors are with the School of Artificial Intelligence and Robotics, Hunan University, China.}%
\thanks{$^{2}$The authors are also with the National Engineering Research Center of Robot Visual Perception and Control Technology, Hunan University, China.}%
\thanks{$^{3}$The authors are with the College of Mechanical and Vehicle Engineering, Hunan University, China.}%
}

\usepackage{placeins}
\usepackage{bbm}
\usepackage{etoolbox}
\usepackage{multicol}
\usepackage{graphics} 
\usepackage{times} 
\usepackage{amsmath} 
\usepackage{amssymb}  
\usepackage{svg}
\usepackage{mathrsfs}
\usepackage{amsfonts}
\usepackage{wrapfig}
\usepackage{amsmath}

\usepackage{kantlipsum}
\usepackage{subcaption}
\usepackage{marvosym}

\usepackage{graphicx}
\usepackage{float}
\usepackage{ctable}
\usepackage{cuted}
\usepackage{colortbl}
\usepackage{multirow}
\usepackage{wrapfig}
\usepackage[misc,geometry]{ifsym}
\usepackage{algorithm}
\usepackage{algpseudocode}

\usepackage{xcolor}   
\usepackage{pifont}    

\definecolor{rblue}{rgb}{0,0.5,1}
\definecolor{awesome}{rgb}{1.0, 0.13, 0.32}
\definecolor{hollywoodcerise}{rgb}{0.96, 0.0, 0.63}
\definecolor{lasallegreen}{rgb}{0.03, 0.47, 0.19}
\definecolor{hanpurple}{rgb}{0.32, 0.09, 0.98}
\definecolor{green(pigment)}{rgb}{0.0, 0.65, 0.31}

\usepackage[pagebackref=false,breaklinks=true,colorlinks,bookmarks=false]{hyperref}
\hypersetup{colorlinks=true,linkcolor={red},citecolor={hanpurple},urlcolor={magenta}}

\usepackage{bm}
\usepackage{caption}
\usepackage{setspace}

\usepackage{enumitem}
\usepackage{threeparttable}
\newcommand{\TODO}[1][]{\textcolor{red}{\bf [TODO]}}
\setlist[enumerate,1]{itemsep=3pt}

\definecolor{formalgreen}{rgb}{0.1, 0.7, 0.1}  
\definecolor{formalred}{rgb}{0.9, 0.2, 0.2}  

\newcommand{\cmark}{\textcolor{formalgreen}{\checkmark}}  
\newcommand{\xmark}{\textcolor{formalred}{\ding{55}}}           
\definecolor{rblue}{rgb}{0,0.5,1}

\newcommand{\change}[1]{{\textcolor{black}{#1}}}
\newcommand{\changed}[1]{\textcolor{black}{#1}}
\usepackage{xcolor}

\begin{document}

\maketitle
\thispagestyle{empty}
\pagestyle{empty}

\begin{abstract}
\input{tex/0-abs}
\end{abstract}
\section{Introduction}
\input{tex/1-intro}

\section{Related Work}
\input{tex/2-related_work}

\section{Method\label{sec:method}}
\input{tex/3-method}

\section{Experiments}
\input{tex/4-exp}

\section{Conclusion}

\input{tex/5-conclusion}

{\small
\bibliographystyle{IEEEtran}
\bibliography{references}
}

\end{document}

%% file: tex/0-abs.tex
Dexterous grasp datasets are vital for embodied intelligence, but mostly emphasize grasp stability, ignoring functional grasps needed for tasks like opening bottle caps or holding cup handles. Most rely on bulky, costly, and hard-to-control high-DOF Shadow Hands. Inspired by the human hand’s underactuated mechanism, we establish UniFucGrasp, a universal functional grasp annotation strategy and dataset for multiple dexterous hand types. Based on biomimicry, it maps natural human motions to diverse hand structures and uses geometry-based force closure to ensure functional, stable, human-like grasps. This method supports low-cost, efficient collection of diverse, high-quality functional grasps. Finally, we establish the first multi-hand functional grasp dataset and provide a synthesis model to validate its effectiveness. Experiments on the UFG dataset, IsaacSim, and complex robotic tasks show that our method improves functional manipulation accuracy and grasp stability, demonstrates improved adaptability across multiple robotic hands, helping to alleviate annotation cost and generalization challenges in dexterous grasping. The project page is at~\href{https://haochen611.github.io/UFG}{https://haochen611.github.io/UFG}.

%% file: tex/1-intro.tex
Functional dexterous grasping has attracted increasing attention due to its critical role in enabling robots to perform complex tasks such as tool use and human-like daily activities~\cite{Brahmbhatt_Handa_Hays_Fox_2019, Zhu_Wu_Lin_Sun_2021, Zhang_Hang_Zhu_Lin_Wu_Peng_Tian_Sun_2023, liu2024realdex}.
Unlike conventional stable grasps, functional grasps require not only secure holding but also task-specific coordination between the hand and the object~\cite{Brahmbhatt_Handa_Hays_Fox_2019}. 
For example, a hammer is typically grasped by the handle during use, but may be held by the head when being handed to another person. 
These nuanced differences highlight the need for fine-grained, semantically meaningful hand-object pose alignment.

Despite their importance, the development of functional dexterous grasping has long been hindered by the lack of large-scale annotated datasets. 
This is mainly due to the high Degrees of Freedom (DoF) in dexterous hands, which makes the annotation process extremely costly and complex. 
Early studies~\cite{DBLP:conf/iros/LiuP0GM19, DBLP:journals/ral/WeiLWLLLZ22, DBLP:conf/icra/WangZCXLLW23, ye2025dex1b} have focused primarily on grasp stability, overlooking the role of task-specific semantic alignment in manipulation.

\begin{figure}[t]
    \centering
    \includegraphics[width=1.0\linewidth]{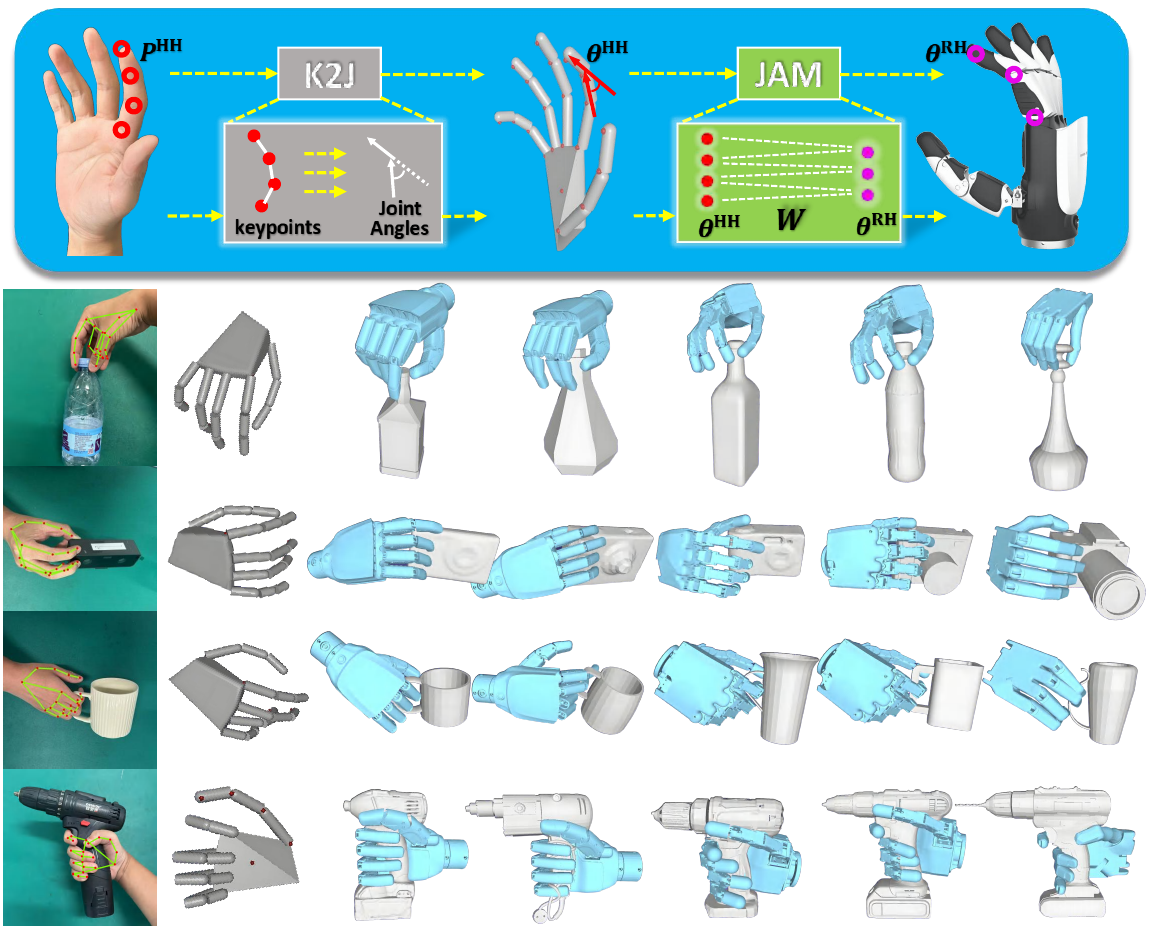}
    \caption{\change{Mapping natural human motions to an anthropomorphic hand model and unified control across representative robotic hands (ShadowHand, InspireHand), and HnuHand~\cite{10011809}. Visualization of functional grasps from UniFucGrasp dataset, including interactions with \textit{bottle}, \textit{camera}, \textit{mug}, and \textit{drill}.}}
    \label{fig:figure1}
    \vspace{-10pt} 
\end{figure}

In the vision community, a common practice is to use the MANO hand model~\cite{DBLP:conf/cvpr/HampaliROL20, DBLP:conf/cvpr/ChaoYXMHTNWIBKF21, DBLP:conf/iccv/JianLLHL23, DBLP:conf/cvpr/YangL0WX0L22} for synthesizing grasp motions. 
However, due to the absence of physical embodiment, the MANO model must be post-processed to map its output to real robotic hands, limiting its applicability in embodied scenarios.

Zhu~\textit{et al.}~\cite{Zhu_Wu_Lin_Sun_2021} were among the first to propose a functional grasp dataset for dexterous hands. 
Their method used binary encodings to annotate contact relationships between object surfaces and finger joints, but suffered from low efficiency and limited data scale. 
Later, Yang~\textit{et al.}~\cite{10752517} introduced a triplet-based semantic graph linking functional fingers to grasp gestures, enabling human-like behavior synthesis. 
However, their approaches were based on symbolic knowledge encoding and lacked real pose supervision.

\change{Recently, DexVLG~\cite{he2025dexvlg} and DexFuncGrasp~\cite{hang2024dexfuncgrasp} proposed large-scale functional grasp datasets (DexGraspNet 3.0 and DFG), providing valuable resources for training vision-language-action systems. 
Nonetheless, both datasets only support annotation for ShadowHand, a fully-actuated and high-cost robotic hand. 
This restricts their accessibility and hinders generalization to real-world applications due to the expensive hardware requirements.}

This leads to a key question: Can we design a cost-efficient and generalizable annotation method that enables functional dexterous grasping across various hand types, facilitating broader adoption?

In fact, the design of robotic hands is often inspired by human motion coordination principles. 
Fully-actuated hands, such as ShadowHand, replicate independent joint control, while underactuated hands like InspireHand leverage mechanical linkages to simplify control. 
Inspired by this, we propose a novel human-to-robot grasp mapping framework that reformulates human motion transfer as a sparse matrix optimization problem. 
This unified formulation serves as a bridge between human demonstration and diverse dexterous hand architectures (both fully- and under-actuated), \changed{facilitating} efficient and versatile functional grasp annotation.

Specifically, we propose a general mapping function that uses the human hand posture as an intermediary to bridge the structural differences between the human hand and various heterogeneous robotic hands. 
This function explicitly establishes the correspondence of Degrees of Freedom (DoFs) between the human hand and robotic hands through an adjustable mapping matrix \( W \). 
Based on this mapping function, by incorporating the degrees of freedom of the target robotic hand \( d_{\mathrm{RH}} \), the degrees of freedom of the human hand \( d_{\mathrm{HH}} \), and their coupling relationships, the weight parameters in the mapping matrix \( W \) linking the joint angles of different heterogeneous robotic hands can be adjusted, while synchronously tuning the coupling matrix \( J \), directly generating control commands for the corresponding robotic hand. 
Leveraging this design, we achieve structural decoupling in terms of DoF and actuation, and unified mapping modeling across diverse hands, eliminating the dependency on specific mechanical architectures, thereby facilitating efficient and precise adaptation of multiple heterogeneous robotic hands through the human hand as an intermediary.

Building on a general and efficient human-to-robot pose mapping method and using MuJoCo~\cite{todorov2012mujoco}, we constructed and released UniFucGrasp (see Fig.~\ref{fig:figure1})—a large-scale functional grasp dataset with over $100K$ high-quality annotations across $1,108$ objects from $21$ daily-use categories. Supporting both fully-actuated and under-actuated dexterous hands, including ShadowHand, InspireHand, and HnuHand~\cite{10011809}, the dataset \changed{enables stable grasp transfer on selected representative robotic hands and generalization to unseen instances within some categories, supporting improved cross-platform consistency.} By employing a unified and novel pose mapping strategy, UniFucGrasp accurately replicates human hand motions on diverse robotic hands, providing stable, consistent functional grasp representations to support task-driven dexterous manipulation research.

In addition, \changed{we present a functional gesture generation model conditioned on point clouds for multiple dexterous hands.} The backbone CVAE~\cite{Sohn_Yan_Lee_2015} learns shared grasp latent features, which a classification head maps to each hand’s DOF space, \changed{facilitating} a generalizable grasping strategy. Experiments in both IsaacSim~\cite{monteiro2019simulating} and real-world scenarios demonstrate significant improvements in functional manipulation accuracy and grasp stability, as well as efficient generalization across different robotic hands on identical tools and tasks.

Our main contributions are summarized as follows:
\begin{itemize}
    \item We propose an annotation strategy and adopt a general, efficient human-to-robot pose mapping method that, using sparse matrix optimization and force-closure analysis, enables stable and reliable functional grasp transfer across diverse dexterous hands, effectively bridging structural and actuation differences.
    
    \item We construct the large-scale \textbf{UniFucGrasp} dataset, containing $1108$ objects from $21$ categories and over $100K$ functional grasp pose annotations, supporting dexterous hands with diverse structures and actuation types, including both fully- and under-actuated designs.
    
    \item \changed{We propose a functional gesture generation model conditioned on hand–object point clouds. Leveraging human prior annotations and joint training across multiple dexterous hands, it achieves unified functional grasping generation with improved precision, stability, and generalization, verified in both simulation and real-world experiments.}
\end{itemize}

%% file: tex/2-related_work.tex
\begin{table*}[!t]
    \caption{Comparison of dexterous grasp datasets (F: Fully-actuated, U: Under-actuated).}
    \resizebox{\textwidth}{!}{
        \begin{tabular}{@{}ccccccccc@{}}
            \toprule
            \begin{tabular}[c]{@{}c@{}}Dataset\end{tabular} & \begin{tabular}[c]{@{}c@{}}  Robot Hand\\ Type (F/U)\end{tabular} & Grasp Method & \begin{tabular}[c]{@{}c@{}}Observations \end{tabular} & \begin{tabular}[c]{@{}c@{}}Sim./Real \end{tabular} & \begin{tabular}[c]{@{}c@{}}Grasps \end{tabular} & \begin{tabular}[c]{@{}c@{}}Obj. (Cat.) \end{tabular} & \begin{tabular}[c]{@{}c@{}}Collection Method\end{tabular} & \begin{tabular}[c]{@{}c@{}}Data Generalization Across\\ Diverse Hands\end{tabular} \\
            \midrule
            HO3D~\cite{DBLP:conf/cvpr/HampaliROL20} & MANO (F) & Stable & RGBD & Real & 77K & 10 & Estimation & \xmark \\
            DexYCB~\cite{DBLP:conf/cvpr/ChaoYXMHTNWIBKF21} & MANO (F) & Stable & RGBD & Real & 582K & 20 &  Manual Annotation & \xmark \\
            DexGraspNet~\cite{DBLP:conf/icra/WangZCXLLW23} & ShadowHand (F) & Stable & - & Sim & 1.32M & 5355 (133) & Optimization & \cmark \\
            AffordPose~\cite{DBLP:conf/iccv/JianLLHL23} & MANO (F) & Functional & - & Sim & 26k & 641 (13) & Optimization & \xmark \\
            OakInk~\cite{DBLP:conf/cvpr/YangL0WX0L22} & MANO (F) & Functional & RGBD & Sim & 1K & 100 (12) & Optimization & \xmark \\
            Toward human-like grasp~\cite{Zhu_Wu_Lin_Sun_2021} & ShadowHand (F) & Functional & Semantic Knowledge & Real & - & 129 (18) & Manual Annotation & \xmark \\
            F2F \cite{10752517} & InspireHand (U) & Functional & Semantic Knowledge & Real & 14 & 127 (18) & Manual Annotation & \xmark \\
            DexFuncGrasp~\cite{hang2024dexfuncgrasp} & ShadowHand (F) & Functional & RGBD & Real-Sim & 14K & 559 (12) & Optimization & \xmark \\
            \textbf{UniFucGrasp (Ours)} & 
            \begin{tabular}[c]{@{}c@{}} \textbf{ShadowHand (F), InspireHand (U),} \\ \textbf{HnuHand (U)} \end{tabular} & 
            \textbf{Stable, Functional} & 
            RGBD & 
            Real-Sim & 
            \textbf{100K} & 
            \textbf{1108 (21)} & 
            \textbf{Human Hand Mapping} & 
            \cmark \\
            \bottomrule
        \end{tabular}
    }
    \label{tab:relatedgrasp}
    \vspace{-5pt}
\end{table*}

\subsection{Dexterous Robot Grasp Datasets}
Existing dexterous hand grasping datasets~\cite{DBLP:conf/iros/LiuP0GM19} primarily focus on grasp stability, typically by directly sampling contact points on the object surface and evaluating grasp robustness using the GraspIt platform~\cite{Miller_Allen_2000}. 
One approach~\cite{Liu_Liu_Jiao_Zhu_Zhu_2022} is to use the force closure criterion as the optimization objective to improve the stability and quality of the grasp. 
Another approach~\cite{DBLP:conf/icra/WangZCXLLW23} focuses on sampling better initial grasp poses to further optimize the final target hand configurations. Although these methods have improved grasp performance to some extent, their reliance on simple grasping strategies and datasets still limits their ability to scale toward functional manipulation tasks.
Zhu~\textit{et al.}~\cite{Zhu_Wu_Lin_Sun_2021} proposed a functional grasping dataset for dexterous hands. They used manually annotated binary codes to represent the contact relationships between the hand and the object, but this method is inefficient and lacks pose data from real robotic platforms. 
Although recent research~\cite{hang2024dexfuncgrasp} introduced a method for collecting functional grasping data from human hand motions in real time, it relies on deep learning models specifically trained for the shadow hand~\cite{Li_Ma_Liang_Görner_Ruppel_Fang_Sun_Zhang_2018}, resulting in significant hand-type dependency and limited generalization performance to other types of robotic hands. 
In addition, the lack of systematic evaluation of grasp stability leads to unreliable generated gestures, making it difficult to effectively support complex grasping and manipulation tasks across dexterous robotic hands.

\change{Unlike existing datasets shown in Table~\ref{tab:relatedgrasp}, this work establishes a unified functional grasping dataset for diverse dexterous hands, integrating grasp stability and functionality, with grasp data collected via human hand mapping, to alleviate the limitations of existing datasets in generalization capability and real-pose acquisition, thereby advancing complex dexterous manipulation tasks.}

\subsection{Human-to-Robotic Hand Motion Mapping}
One of the key challenges in constructing dexterous hand grasp datasets is efficiently mapping natural human hand motions to various robotic hands. Existing methods mainly include joint mapping for power grasps~\cite{Kobayashi_Kitabayashi_Nakamoto_Kojima_Fukui_Imamura_Maeda_2012, liu2019high, Liarokapis2013Telemanipulation}, fingertip mapping for precision grasps~\cite{rohling1993optimized, cui2014optimization}, and pose mapping for conveying functional intent~\cite{Meeker_Rasmussen_Ciocarlie_2018}. Additionally, dimensionality reduction strategies based on postural synergies have been applied to plan and control fully actuated robotic hands~\cite{ciocarlie2007dimensionality, ficuciello2012planning, palli2014dexmart}. However, these methods are typically designed for a single type of robotic hand and overly rely on a single mapping paradigm, which limits their ability to scale in terms of generalizability and functional effectiveness. To address this, we propose a unified mapping strategy that models skeletal keypoints, joint angles, and actuation values as a sparse matrix, preserving natural human motion patterns \changed{while aiming for compatibility with various dexterous hands and alleviating dependence on large-scale training data.} Based on this, we constructed the Unified Functional Grasp (UFG) dataset, providing annotated functional grasp poses and object category labels for multiple robotic hands.

%% file: tex/3-method.tex
\begin{figure*}[t] \centering
    \includegraphics[width=0.98 \linewidth]{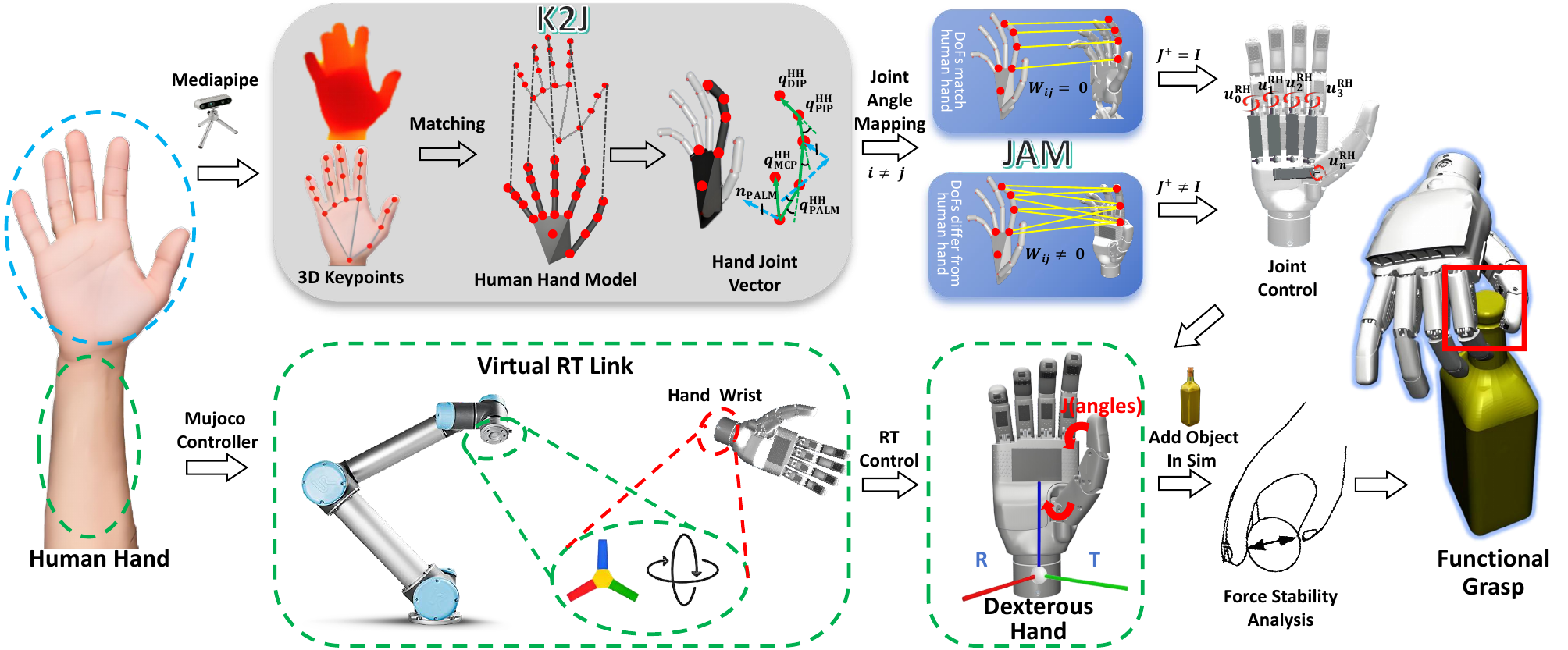}
    \caption{\change{Illustration of our annotation strategy for transferring human motions to robotic hands, with the left showing capture, the top depicting gesture mapping, the bottom showing pose collection, and the right displaying the resulting functional grasps.}} \label{fig:pipeline}
    \vspace{-5pt}
\end
{figure*}

In this work, the goal is to generate reliable, functional grasp poses for diverse robotic hands via Human-Hand mapping. Given the object mesh, robot hand URDF, and $21$ 3D keypoints from an RGB-D camera, we build a unified action mapping represented by pose (rotation \(R\), translation \(T\)) and joint angles \(J\). 
We aim to provide an efficient pipeline to ensure reliable grasp generation and strong generalization for dexterous hands. 
An overview of our method is shown in Fig.~\ref{fig:pipeline}. 

\textbf{Method Overview.}
First, we extract human hand skeletal keypoints and build a kinematic model, designing a keypoint-to-joint-angle conversion network (K2J) (see Sec.~\ref{subsec:pretrain}) to faithfully replicate human hand motions. The K2J module uses MediaPipe~\cite{Lugaresi_Tang_Nash_McClanahan_Uboweja_Hays_Zhang_Chang_Yong_Lee_etal._2019} for keypoint detection, transforms them to the camera coordinate system, and maps them onto a biomimetic model closely matching human hand proportions, accurately reconstructing hand kinematics. Next, we represent the mapping from the human hand to various robotic hands (JAM) (see Sec.~\ref{subsec:arch}) as a sparse matrix optimization, capturing anthropomorphic grasp poses. 
Using this mapping, we derive robotic joint angles and convert them from joint space to actuation space via a standardized method (see Sec.~\ref{sec:graspconfig}), enabling stable control of the simulated hand and producing high-quality functional grasps. 
Finally, based on this annotation method, we build a multi-hand functional grasp dataset and validate its effectiveness within the functional gesture generation framework (see Sec.~\ref{subsec:fucgraspgener}).

\subsection{Human-Hand Kinematic Modeling} \label{subsec:pretrain}
To enable functional grasping, robots must understand the human-hand structure and motion. 
We decompose this into: (1) anthropomorphic hand model alignment, abstracting the hand into a biomimetic model to determine joint relative positions; (2) joint angle estimation, constructing a kinematic model for high-fidelity motion reconstruction.

Using MediaPipe~\cite{Lugaresi_Tang_Nash_McClanahan_Uboweja_Hays_Zhang_Chang_Yong_Lee_etal._2019}, we detect $21$ 2D hand keypoints $\{\mathbf{k}_i\}_{i=1}^{21}$ from RGB images, then project them into 3D camera-frame points $\{\mathbf{p}_i\}_{i=1}^{21}$ with depth maps and camera intrinsics $\mathbf{K}_{\text{int}}$. Keypoints are registered onto the biomimetic hand model, replacing the wrist keypoint with the palm center for stability and accurate palm normal vector estimation. 
As shown in Fig.~\ref{fig:pipeline}, the palm normal vector $\mathbf{n}_{\text{PALM}}$ is computed by the cross product of vectors formed by the ring finger base, index finger base, and palm center, using the index finger as an example:
\begin{equation}
\mathbf{n}_{\text{Palm}} = (\mathbf{p}_{\text{Ring}} - \mathbf{p}_{\text{Palm}}) \times (\mathbf{p}_{\text{Index}} - \mathbf{p}_{\text{Palm}}).
\end{equation}

Next, for the \(i\)-th finger, we define the vector from joint \(n\) to joint \(n+1\) as the \(n\)-th joint vector of the finger, denoted as \(\mathbf{q}_{i,n}^{\mathrm{HH}}\). This MCP joint vector is projected onto the palm plane defined by the palm normal vector \(\mathbf{n}_{\mathrm{Palm}}\), and the resulting projection vector is denoted as \(\mathbf{n}_{i,1}^{\mathrm{HH}}\).
The abduction-adduction angle of the \(i\)-th finger is the angle between the palm-to-MCP vector \(\mathbf{q}^{\mathrm{HH}}_{i,0}\) and The projection vector \(\mathbf{n}^{\mathrm{HH}}_{i,1}\), given by:
\begin{equation}
\theta^{\mathrm{HH}}_{i,\text{abd}} = \arccos \left( \frac{ \mathbf{q}_{i,0}^{\mathrm{HH}} \cdot \mathbf{n}^{\mathrm{HH}}_{i,1} }{ \| \mathbf{q}_{i,0}^{\mathrm{HH}} \| \cdot \| \mathbf{n}^{\mathrm{HH}}_{i,1} \| } \right).
\end{equation}

The flexion-extension angle $\theta_{i,\text{flex}}^{\mathrm{HH}}$ is defined as the angle between the current joint vector and the reverse extension of the adjacent joint vector:
\begin{equation}
\theta_{i,\text{flex}}^{\mathrm{HH}} = \arccos \left( \frac{ \mathbf{q}_{i,n}^{\mathrm{HH}} \cdot \mathbf{q}_{i,(n+1)}^{\mathrm{HH}} }{ \| \mathbf{q}_{i,n}^{\mathrm{HH}} \| \cdot \| \mathbf{q}_{i,(n+1)}^{\mathrm{HH}} \| } \right).
\end{equation}

\changed{The overall joint angle prediction error is calculated by:
\begin{equation}
\mathit{E} = \sqrt{
\frac{1}{N} \sum_{i=0}^{f} \sum_{n=0}^{d_i} \left( \hat{\theta}_{i,n}^{\mathrm{HH}} - \theta_{i,n}^{\mathrm{HH}} \right)^2.
}
\label{eq:joint_error}
\end{equation}
where \(N\) denotes the total number of data samples, \(f\) is the total number of fingers, and \(d_i\) represents the number of joints in the \(i\)-th finger. Here, \(\hat{\theta}_{i,n}^{\mathrm{HH}}\) and \(\theta_{i,n}^{\mathrm{HH}}\) denote the predicted and ground-truth joint angles of the \(n\)-th joint in the \(i\)-th finger of the Human-Hand, respectively.}

\begin{figure}[t!] \centering
\includegraphics[width=1.00\linewidth]{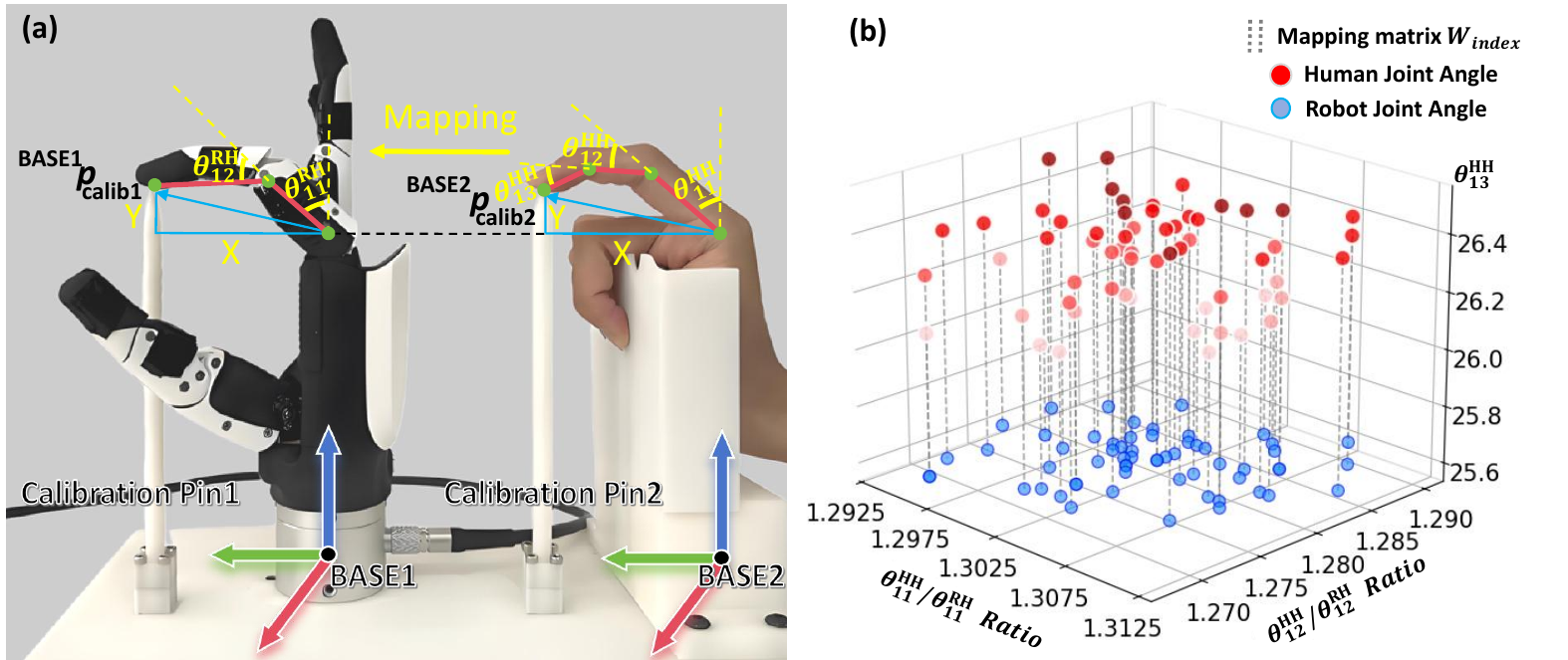}
    \caption{Flowchart of data acquisition for mapping joint angles from the human index finger to the robotic index finger.}
    \vspace{-10pt}
    \label{fig:test}
\end{figure}

\subsection{Human-Hand Mapping Representation} \label{subsec:arch}
Given the human hand grasping pose represented by the joint angles
\(\theta \in \mathbb{R}^{20 \times 1}\), our objective is to construct a general mapping function that enables accurate replication of the Human-Hand posture on the robotic platform. Specifically, the \(n\)-th joint angle of the \(i\)-th finger in the Human-Hand is denoted as \(\theta_{i,n}^{\mathrm{HH}}\), and the corresponding joint angle in the robotic hand is denoted as \(\theta_{i,n}^{\mathrm{RH}}\). 
The mapping relationship is formulated as:
\begin{equation}
\theta_{i,n}^{\mathrm{RH}} = W \theta_{i,n}^{\mathrm{HH}} + \epsilon,
\end{equation}
where \(W\) is the mapping matrix and \(\epsilon\) is the error term capturing possible deviations.
The dimension of the mapping matrix \( W \) depends on the relationship between the degrees of freedom of the robotic hand \( d_{\mathrm{RH}} \) and the Human-Hand \( d_{\mathrm{HH}} \), specifically expressed as:
\begin{equation}
\dim(W) =
\begin{cases}
\mathbb{R}^{d_{\mathrm{RH}} \times d_{\mathrm{HH}}} = \mathbb{R}^{d_{\mathrm{HH}} \times d_{\mathrm{HH}}}, & \text{if } d_{\mathrm{RH}} = d_{\mathrm{HH}}, \\
\mathbb{R}^{d_{\mathrm{RH}} \times d_{\mathrm{HH}}}, & \text{if } d_{\mathrm{RH}} \neq d_{\mathrm{HH}}.
\end{cases}
\end{equation}
When \( d_{\mathrm{RH}} = d_{\mathrm{HH}} \), the mapping matrix \( W \) is square, enabling one-to-one joint correspondence. When \( d_{\mathrm{RH}} \neq d_{\mathrm{HH}} \), \( W \) becomes non-square, performing compression or expansion to adapt the Human-Hand joint space to the robotic hand.

For the ShadowHand (\( d_{\mathrm{RH}} = d_{\mathrm{HH}} \)), \( W \) is a diagonal matrix that scales joint angles based on size differences. For the InspireHand (\( d_{\mathrm{RH}} = 12 \), \( d_{\mathrm{HH}} = 20 \)), Human-Hand joints \( \theta^{\mathrm{HH}} \in \mathbb{R}^{20 \times 1} \) are mapped to robotic hand joints \( \theta^{\mathrm{RH}} \in \mathbb{R}^{12 \times 1} \) via \( W \in \mathbb{R}^{12 \times 20} \).

\begin{equation}
\theta^{\mathrm{HH}} \in \mathbb{R}^{20 \times 1} \xrightarrow{W} \theta^{\mathrm{RH}} \in \mathbb{R}^{12 \times 1}.
\end{equation}
The mapping matrix \( W \) can be decomposed into submatrices corresponding to each finger:
\begin{equation}
W = 
\begin{bmatrix}
W_{\mathrm{Thumb}} & W_{\mathrm{Index}} & W_{\mathrm{Middle}} & W_{\mathrm{Ring}} & W_{\mathrm{Little}}
\end{bmatrix}^T,
\end{equation}
where each submatrix  \changed{\( W_{\text{Thumb}} \), \( W_{\text{Index}} \), \( W_{\text{Middle}} \), \( W_{\text{Ring}} \), and \( W_{\text{Little}} \)} maps the joint angles of a specific finger from the Human-Hand to the corresponding finger on the robotic hand.
Taking the index finger as an example, since the InspireHand lacks the abduction degree of freedom and has one fewer flexion/extension DoF compared to the Human-Hand, it has $2$ DoFs instead of $4$. 
Thus, the two joint angles of the InspireHand index finger, $\theta^{\mathrm{RH}}_{11}$ and $\theta^{\mathrm{RH}}_{12}$, are linearly mapped from the three joint angles of the Human-Hand index finger, $\theta^{\mathrm{HH}}_{11}$, $\theta^{\mathrm{HH}}_{12}$, and $\theta^{\mathrm{HH}}_{13}$, through the index finger mapping submatrix $W_{\text{Index}}$, as follows:
\begin{equation}
\begin{bmatrix}
\theta^{\mathrm{RH}}_{11} \
\theta^{\mathrm{RH}}_{12}
\end{bmatrix}^{\mathrm{T}}
= W_{\text{Index}}
\begin{bmatrix}
\theta^{\mathrm{HH}}_{11} & \theta^{\mathrm{HH}}_{12} & \theta^{\mathrm{HH}}_{13}
\end{bmatrix}^{\mathrm{T}},
\end{equation}
\changed{and the mapping submatrix \( W_{\mathrm{Index}} \in \mathbb{R}^{2 \times 3} \) is defined and optimized as:}
\begin{equation}
W_{\mathrm{Index}} =
\begin{bmatrix}
\alpha & \beta & \gamma \\
\delta & \varepsilon & \zeta
\end{bmatrix},
\quad
\changed{\min_{W_{\mathrm{Index}}}
\sum_{i=1}^{N}
\left\|
W_{\mathrm{Index}} \, \boldsymbol{\theta}_{i}^{HH}
-
\boldsymbol{\theta}_{i}^{RH}
\right\|_2^2.}
\label{eq:w_index_optimization}
\end{equation}
\changed{The parameters $\alpha, \beta, \gamma, \delta, \varepsilon, \zeta$ are obtained through a fingertip alignment-based mapping optimization method. By synchronously collecting motion data of the human and robotic hands under consistent fingertip contact and pressing postures(Fig.~\ref{fig:test}(a)), a linear mapping $W_{\text{Index}}$ is established between their joint angle spaces(Fig.~\ref{fig:test}(b)). Finally, the parameters are optimized via least squares, constrained by fingertip position alignment.}
\changed{The mapping must satisfy both joint feasibility and fingertip position consistency:
\begin{align}
\theta^{RH}_{i,\min} &\le \theta^{RH}_i \le \theta^{RH}_{i,\max}, \label{eq:constrain}\\
f^{HH}(\theta^{HH}_i) &= f^{RH}(\theta^{RH}_i) = {}^{\mathrm{BASE}}p_{\mathrm{calib}}. \label{eq:calib}
\end{align}
where $f^{HH}(\cdot)$ and $f^{RH}(\cdot)$ are the forward kinematics functions of the human and robotic hands, respectively, mapping joint angles to fingertip positions; ${}^{\mathrm{BASE}}p_{\mathrm{calib}}$ represents the fingertip position expressed in the base coordinate frame(Fig.~\ref{fig:test}(a)).
}

\begin{figure}[!t]
    \centering
    \includegraphics[width=\linewidth]{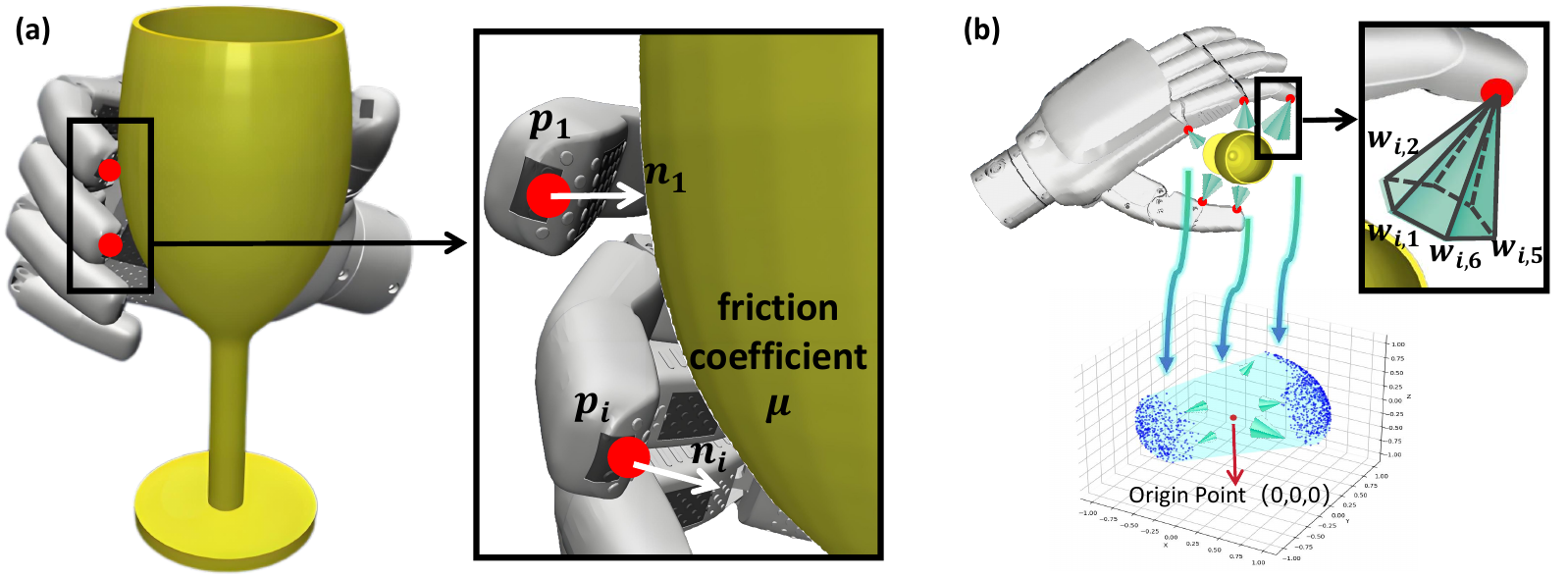}
    \caption{\changed{\change{For each contact $i$ $(\mathbf{n}_i, \mathbf{p}_i)$, we show its six friction cone directions $\mathbf{w}_{i,j}$ and the reduced wrench spaces (blue regions).}}
}
    \vspace{-15pt}
    \label{fig:fig5}
\end{figure}
\subsection{Functional Dexterous Hand Control via RTJ Mapping}\label{sec:graspconfig}
Six virtual links connect the dexterous hand base to the world frame, enabling explicit rotation and translation control. The simulation provides real-time hand pose feedback (quaternions and position) relative to the object. Accurate control requires mapping joint space to actuator space, accounting for motor inputs and constraints. This mapping is direct for fully actuated hands and accounts for coupling in underactuated hands. To unify both, the joint-to-actuator mapping is defined as:
\begin{equation}
u^{\mathrm{RH}}_{i,n} = J^{+} \theta^{\mathrm{RH}}_{i,n},
\label{eq:jacobian}
\end{equation}
where \( J^{+} \) is the generalized inverse of the mapping matrix \( J \), which is commonly computed as the Moore-Penrose~\cite{Barata_Hussein_2012} pseudoinverse:
\begin{equation}
J^{+} = (J^\top J)^{-1} J^\top.
\label{eq:jacobian_pseudoinverse}
\end{equation}

\changed{For the underactuated InspireHand, the joint coupling matrix \(J \in \mathbb{R}^{12 \times 6}\) was obtained via manual measurements, and its generalized inverse \(J^+\) was computed to map joint space to actuator space.}

\changed{After actively mapping and collecting gestures, grasp performance is post-processed and evaluated using geometry-based force-closure analysis~\cite{ciocarlie2007dimensionality,Ciocarlie_Allen_2009}. We record hand–object collision points via open interfaces~\cite{todorov2012mujoco} to define contacts, and approximate each friction cone with six rays to model contact forces.} As shown in Fig.~\ref{fig:fig5}(a), from the grasp contact points, we discretize each friction cone to approximate feasible contact forces. For each contact point $\mathbf{p}_i \in \mathbb{R}^3$, with normal $\mathbf{n}_i \in \mathbb{R}^3$ and friction coefficient $\mu$, the friction cone half-angle is computed as:
\begin{equation}
\theta = \arctan(\mu).
\end{equation}
Given a vector \( \mathbf{r} \) that is not parallel to the normal vector \( \mathbf{n}_i \), we construct two unit vectors \( \mathbf{t}_1 \) and \( \mathbf{t}_2 \) orthogonal to \( \mathbf{n}_i \) via the cross product:
\begin{equation}
\mathbf{t}_1 = \frac{\mathbf{n}_i \times \mathbf{r}}{\| \mathbf{n}_i \times \mathbf{r} \|}, \quad \mathbf{t}_2 = \mathbf{n}_i \times \mathbf{t}_1.
\end{equation}
Based on these, the \( j \)-th approximate friction cone direction is generated as:
\begin{equation}
\mathbf{w}_{i,j} = \cos(\theta) \mathbf{n}_i + \sin(\theta) \big( \cos(\phi_j) \mathbf{t}_1 + \sin(\phi_j) \mathbf{t}_2 \big), \label{eq:wij}
\end{equation} 
where \( \phi_j = \frac{2 \pi (j - 1)}{6} \) for \( j = 1, 2, \ldots, 6 \).

For \( n \) contact points, each with $6$ directions, the wrench and the grasp matrix \(\mathbf{G}\) are computed as follows, where \( i=1,\ldots,n \) and \( j=1,\ldots,6 \):
\begin{equation}
\mathbf{wrench}_{i,j} = \begin{bmatrix} 
\mathbf{w}_{i,j} \\ 
\mathbf{p}_i \times \mathbf{w}_{i,j} 
\end{bmatrix} \in \mathbb{R}^6, 
\end{equation}
\begin{equation}
\mathbf{G} = [ \mathbf{wrench}_{i,j} ] \in \mathbb{R}^{6 \times 6n}.
\end{equation}
where \( i \) and \( j \) iterate over contact points and directions, respectively. Then, as shown in Fig.~\ref{fig:fig5}(b), the force-closure condition is checked by determining whether the origin lies inside the convex hull of the wrench vectors.

\begin{figure}[!t]
    \centering
    \includegraphics[width=\linewidth]{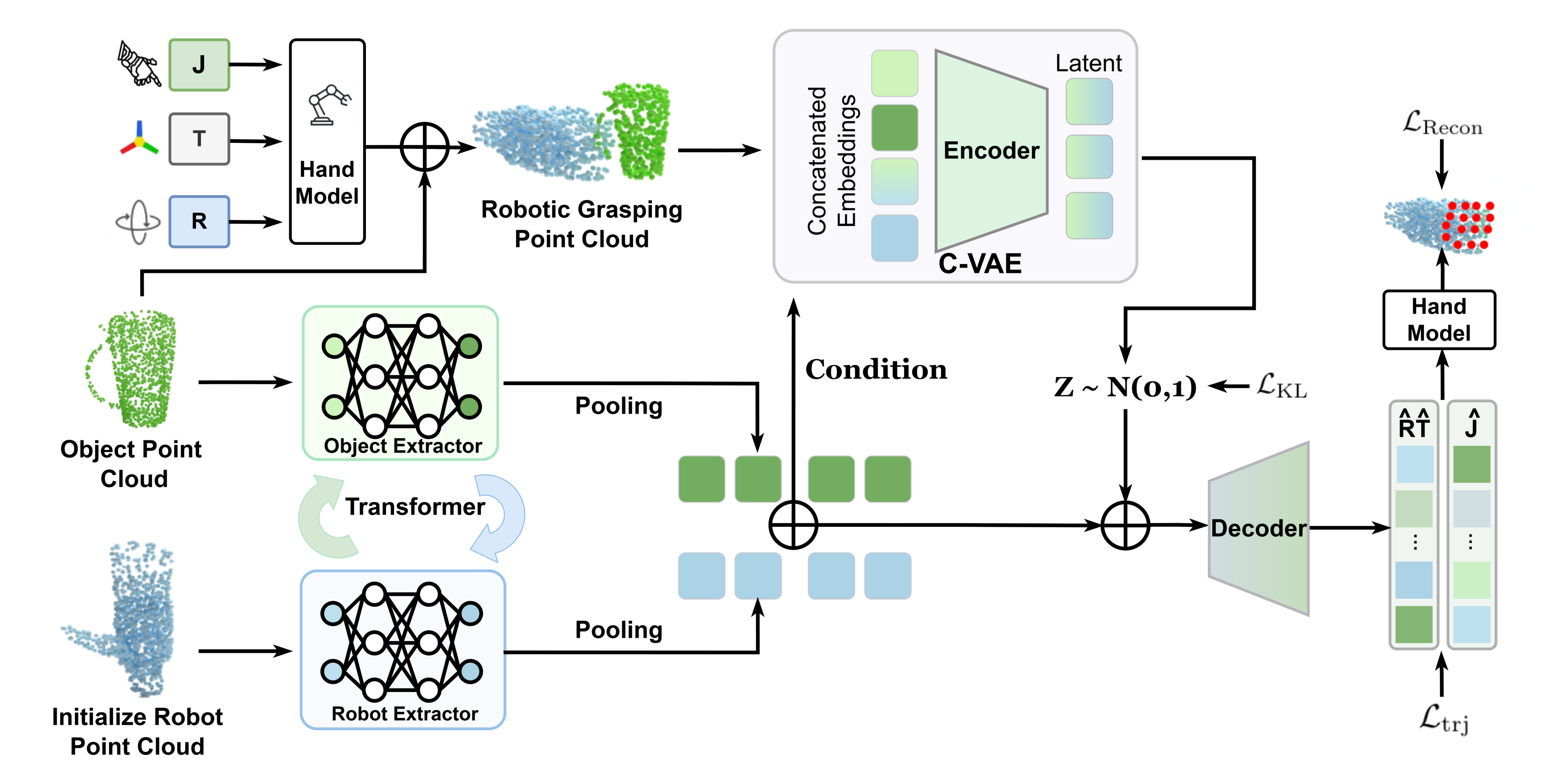}
    \caption{\change{Functional gesture generation for diverse robotic hands, with point cloud feature extraction, CVAE~\cite{Sohn_Yan_Lee_2015} latent encoding and sampling, and MLP decoder for gesture parameters.}}
    \vspace{-15pt}
    \label{fig:model}
\end{figure}

\begin{figure*}[!t]
    \centering
    \includegraphics[width=\textwidth]{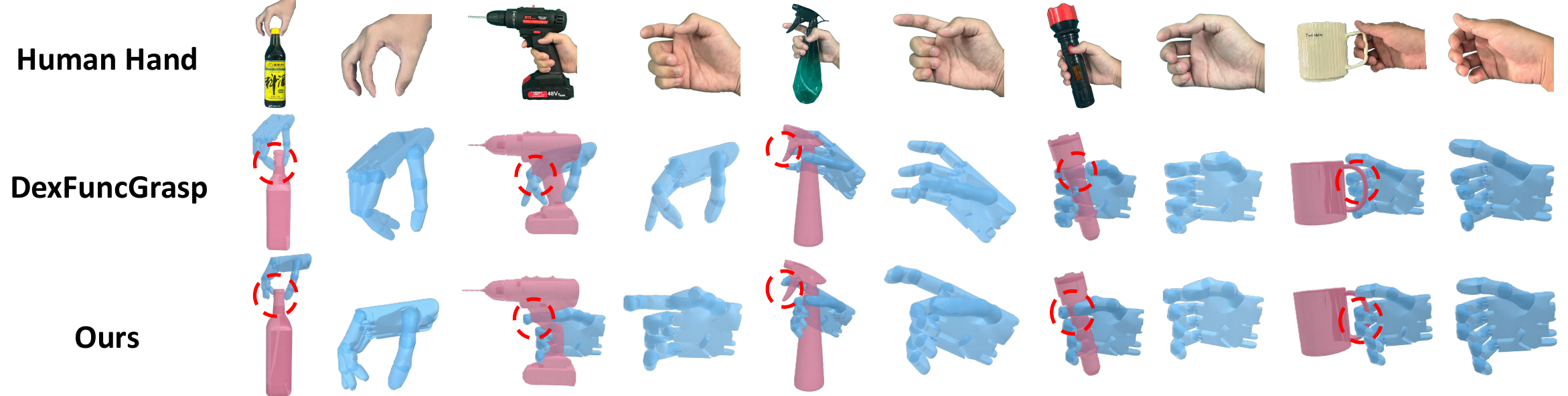}  
    \caption{Qualitative comparison with the state-of-the-art method DFG~\cite{hang2024dexfuncgrasp} on functional grasping shows that our method better captures human hand poses and more consistently aligns with functional regions across diverse manipulation tools.}
    \vspace{-5pt}
    \label{fig:comparison}
\end{figure*}
\subsection{Functional Grasp Generation} \label{subsec:fucgraspgener}
\textbf{Functional Grasp Synthesis Model:} 
Both stable and functional grasping fundamentally depend on accurately predicting the dexterous hand's pose (\(R, T\)) and joint configurations (\(J\)). 
To assess the effectiveness of our dataset for functional grasping, we develop a task-driven, lightweight deep neural network. 
As shown in Fig.~\ref{fig:model}, the designed network takes hand and object point clouds with input dimensions of \(2500 \times 3\) and \(2000 \times 3\), which are separately processed by Robot Extractor and Object Extractor modules based on DGCNN~\cite{Wang_Sun_Liu_Sarma_Bronstein_Solomon_2019}. 
It effectively extracts local geometric features from sparse 3D data by leveraging spatial relationships among neighboring points, providing stable and spatially-aware feature encodings for the subsequent CVAE~\cite{Sohn_Yan_Lee_2015} to generate diverse and plausible hand configurations.

In the designed functional grasp generation network, we adopt a lightweight Transformer-based architecture following DCP~\cite{wang2019deep} for cross-object embedding and cross-modal alignment. 
Fused features from a $4$-head encoder-decoder with $128$ feedforward hidden dimensions are fed into the CVAE~\cite{Sohn_Yan_Lee_2015} encoder. 
A latent vector $z$ is sampled and concatenated with max-pooled hand and object features to form a $260$-dimensional joint representation. 
This vector is fed into the grasp generation network to predict hand rotation (\(r\)), translation (\(t\)), and joint angles (\(j\)).

\textbf{Loss Functions:} \changed{We adopt a compact loss to evaluate functional grasping, where 
$\alpha_1, \alpha_2, \alpha_3$ weight the KL divergence, trajectory, 
and reconstruction terms, respectively:}
\begin{equation}
\mathcal{L} = \alpha_1 \mathcal{L}_{\mathrm{KL}} + \alpha_2 \mathcal{L}_{\mathrm{trj}} + \alpha_3 \mathcal{L}_{\mathrm{Recon}}.
\end{equation}
The KL divergence encourages the latent distribution \(Q(z \mid o, g)\) to align with a standard Gaussian prior \(\mathcal{N}(0, I)\), facilitating structured and continuous latent space learning:
\begin{equation}
\mathcal{L}_{\mathrm{KL}} = - \mathrm{KL}\left( Q(z \mid o, g) \parallel \mathcal{N}(0, I) \right).
\end{equation}
\changed{We first define the L1 loss on grasp parameters: hand rotation ($\hat{r}$), 
translation ($\hat{t}$), and joint angles ($\hat{j}$):}
\begin{equation}
\mathcal{L}_{\mathrm{trj}} = \lambda_1 \lVert \hat{r} - r \rVert_1 + \lambda_2 \lVert \hat{t} - t \rVert_1 + \lambda_3 \lVert \hat{j} - j \rVert_1.
\end{equation}
\begin{equation}
\mathcal{L}_{\mathrm{Recon}} = \sum_{i=0}^{index} \left| p_i^{pred} - p_i^{gt} \right|.
\end{equation}

\changed{In addition, we design a reconstruction loss 
$L_{\text{Recon}}$ in Fig.~\ref{fig:model}, which supervises training by comparing keypoints 
of the predicted grasp gestures—generated from 
$\hat{r}, \hat{t}, \hat{j}$ via the hand model—with the corresponding 
ground-truth gestures. We adopt L1 loss for robustness and stable gradients, and apply it on gesture keypoints to enhance prediction accuracy and model stability.}

%% file: tex/4-exp.tex
\begin{figure}[!t]
    \centering
    \includegraphics[width=\linewidth]{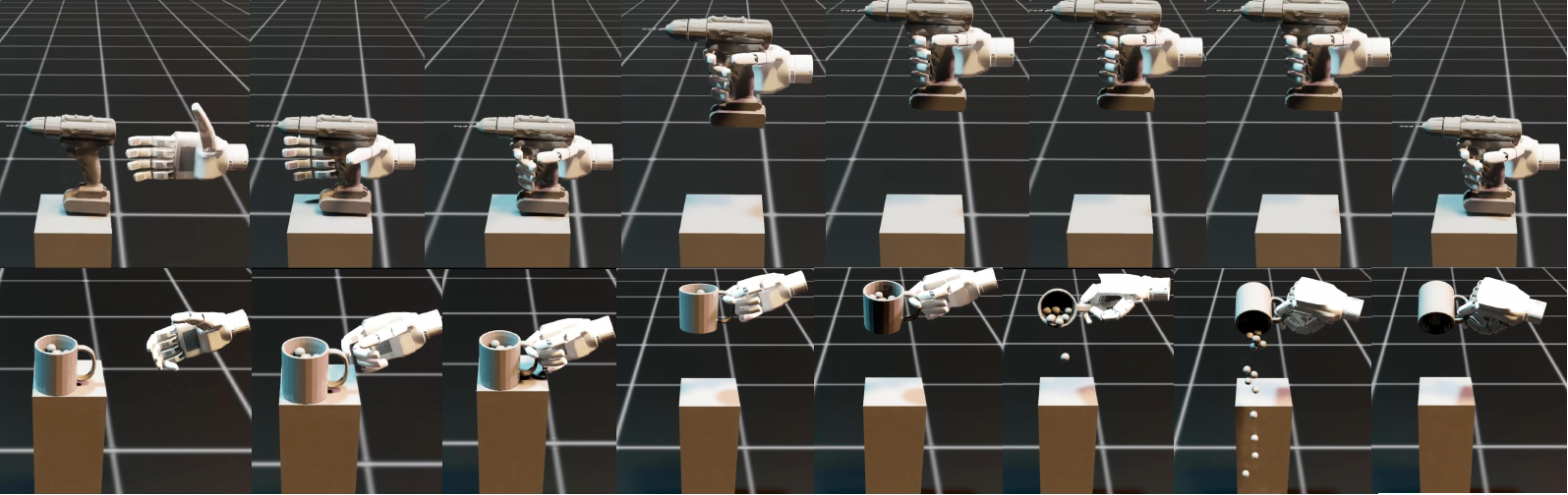}
    \caption{\changed{Simulation results of the method for functional tasks: \textit{Press Button and} \textit{Pour Water}.}}
    \vspace{-15pt}
    \label{fig:simulation}
\end{figure}

\subsection{Experiment Setup}\label{experiment setup}
\textbf{Annotation and Dataset:} 
\changed{We collected 60 sets of human joint angle data from six volunteers using the K2J module and refined the measurements with a bone joint goniometer, constraining joint angle errors according to Eq.~\eqref{eq:joint_error} to improve accuracy. The JAM module was then used to establish fingertip mapping between human and robotic hands. Mapping parameters $W$ were estimated via least-squares optimization to minimize joint angle differences (Eq.~\eqref{eq:w_index_optimization}), with constraints on joint range (Eq.~\eqref{eq:constrain}) and the absolute coordinates of the fingertips and hand base $^{\mathrm{BASE}}p_{\mathrm{calib}}$ (Eq.~\eqref{eq:calib}). Taking the InspireHand as an example, since its thumb shares the same degrees of freedom as the human thumb, a direct mapping was used. For the other fingers, due to differences in degrees of freedom, we employed a calibration experimental setup for fingertip mapping, using the functional index finger as a representative for calibration (Fig~\ref{fig:test}), to achieve a more accurate dimensionality-reducing mapping. 
The mapping matrix \( W_{\mathrm{Index}} \) include \(\alpha {=} 0.3530\), \(\beta {=} 0.4310\), \(\gamma {=} 0.2827\), \(\delta {=} 0.2584\), \(\varepsilon {=} 0.4130\), and \(\zeta {=} {-}0.0018\). To directly translate the mapping results into dexterous hand control and enable efficient annotation of functional grasp postures, we measured and modeled the joint coupling of the InspireHand, constructed the Jacobian matrix \( J \in \mathbb{R}^{12 \times 6} \) mapping joint to actuator space (Eq.~\eqref{eq:jacobian}), and computed its pseudoinverse \( J^+ \) to map joint-level commands to actuator-level signals.}
\begin{figure}[!t]
    \centering
    \includegraphics[width=\linewidth]{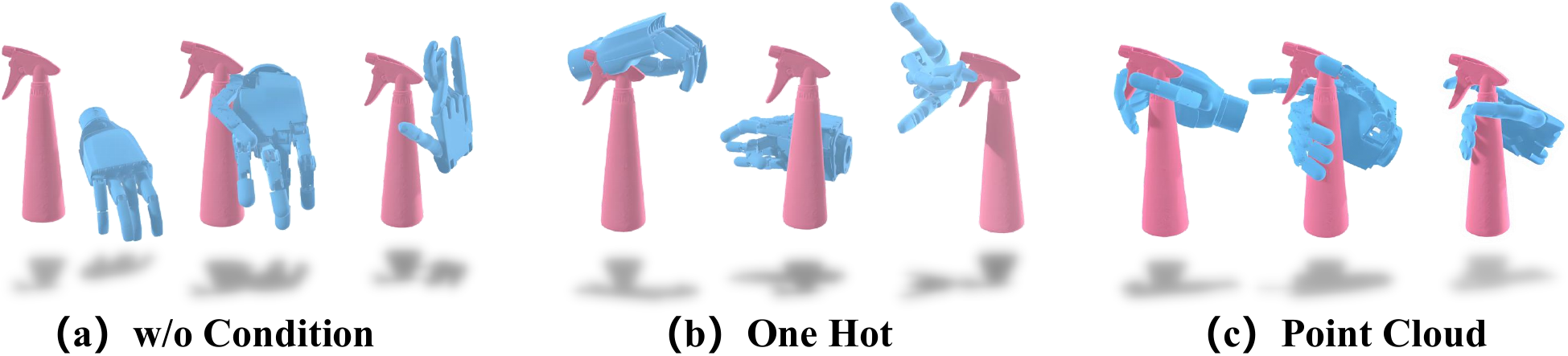}
    \caption{\changed{Qualitative ablation results for functional grasping of \textit{spraybottle} under three different hand-conditioned inputs.}}
    \vspace{-10pt}
    \label{fig:Ablation_view}
\end{figure}
Based on this annotation method, we constructed the UFG dataset in MuJoCo~\cite{todorov2012mujoco} by controlling three dexterous hands via tracked natural hand motions in Fig.~\ref{fig:figure1}. The dataset covers $21$ categories with $1,108$ object instances, each having over $70$ validated functional grasp demonstrations, totaling more than $100K$ annotations. 
\changed{Grasp stability was ensured via force feedback and collision detection, with each contact point defined as the collision between the object and hand, its 3D position and surface normal recorded, and the friction cone discretized into six rays with a coefficient of friction \(\mu {=} 0.5\)~\cite{inouye2012novel, yao2023exploiting}, typical of common materials~\cite{lide1995crc}.} 
\changed{The dataset was split into training and testing sets at an $8.5{:}1.5$ ratio per category, with the test set consisting entirely of unseen objects that are structurally distinct from those in the training set.} 
Our method enhances the DFG dataset~\cite{hang2024dexfuncgrasp}, \change{which helps capture realistic human motion priors and common functional grasp patterns.}

\textbf{Implementation Details:} 
The model employs DGCNN~\cite{Wang_Sun_Liu_Sarma_Bronstein_Solomon_2019} for feature extraction and is trained within a conditional variational autoencoder CVAE~\cite{Sohn_Yan_Lee_2015} using the Adam optimizer, with a learning rate of $0.00001$ over $15$ epochs. Experiments are conducted on two NVIDIA RTX 3090 GPUs. 
To quantitatively evaluate the performance of our dataset and the functional grasp synthesis model, we apply Kullback-Leibler divergence to regularize the latent space and promote structured grasp representations, alongside an L1-based reconstruction loss to measure the accuracy of predicted hand rotation, translation, and joint angles, effectively reflecting the validity and stability of generated functional gestures.
\begin{figure}[!t]
    \centering
    \includegraphics[width=\linewidth]{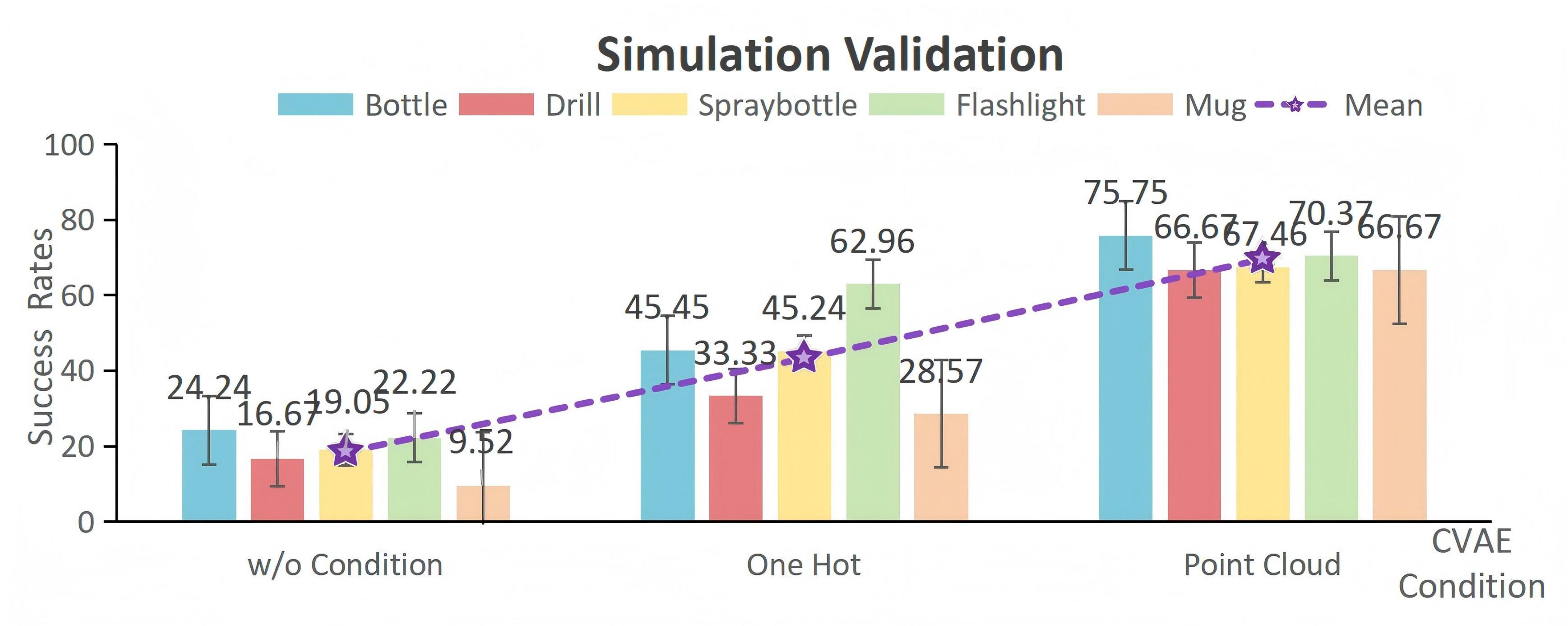}
    \caption{Quantitative results of three hand-conditioned input settings under the same random seed (41, 42, 43).} 
    \vspace{-10pt}
    \label{fig:Ablation}
\end{figure}
\subsection{Comparison of Grasping Performance}\label{Baseline Comparison}

\begin{table}[!t]
\centering
\footnotesize
\caption{\changed{Quantitative analysis of simulation test results across different random seeds for each object category.}}
\resizebox{0.49\textwidth}{!}{
\begin{tabular}{@{}lcccccc@{}}
\toprule
\textbf{Category} & \textbf{SR (DFG~\cite{hang2024dexfuncgrasp})} & \textbf{SR (Seed 41 / 42 / 43)} & \textbf{Mean (Ours) ± Std} & \textbf{Train / Test} \\
\midrule
Bottle       & 0.6862 &  0.7272 / 0.8181 / 0.7272 & \textbf{0.7575 ± 0.0428} & 54 / 11 \\
Drill        & 0.5500 &  0.6250 / 0.7500 / 0.6250 & \textbf{0.6667 ± 0.0589} & 48 / 8  \\
Spraybottle  & 0.5807 &  0.6428 / 0.6667 / 0.7143 & \textbf{0.6746 ± 0.0364} & 85 / 14 \\
Flashlight   & 0.9103 &  0.7777 / 0.6667 / 0.6667 & 0.7037 ± 0.0523 & 44 / 9  \\
Mug          & 0.5462 &  0.5714 / 0.7143 / 0.7143 & \textbf{0.6667 ± 0.0674} & 38 / 7  \\
\textbf{Total} & 0.6552 & 0.6732 / 0.7207 / 0.6934 & \textbf{0.6955 ± 0.0219} & 269 / 48 \\
\bottomrule
\end{tabular}
\vspace{-5pt}
}

\vspace{-15pt}
\label{tab:inspirehand_results}
\end{table}
\textbf{Quantitative Results:} 
\changed{To verify the effect of hand point cloud conditioning in the CVAE framework, we performed an ablation study across different categories with all other components fixed (see Fig.~\ref{fig:Ablation}). Results comparing no condition, One-hot identity encoding, and point cloud conditioning show clear gains in success rates: $19.05\%$, $44.21\%$, and $69.55\%$, respectively—with point cloud significantly improving performance by $57.34\%$, indicating that the point cloud offers richer geometric and spatial cues for generating stable, functional, and unified human-like dexterous grasps.} The model’s predicted grasp poses and joint angles are visualized and executed in IsaacSim~\cite{monteiro2019simulating} simulation, where both hand and object are treated as rigid bodies. 
\changed{A grasp is deemed successful if the object remains held after lifting the hand by $10cm$, and a manipulation is considered successful if the task can be completed under predefined disturbances. All other hyperparameters were kept unchanged, and seeds 41, 42, and 43 were used for testing. Table~\ref{tab:inspirehand_results} shows our method achieves an average success rate of $69.55\%$, about $6.15\%$ higher than DFG~\cite{hang2024dexfuncgrasp}. Although the grasp success rates of the mug and drill are relatively low across all methods due to handle shapes and narrow gaps, our method achieves higher success rates on these objects ($66.67\%$), surpassing $55\%$ (drill) and $54.62\%$ (mug) in DFG~\cite{hang2024dexfuncgrasp}. while also generating gestures that precisely align the index finger with the drill button or execute the mug pouring action, ensuring successful functional manipulations. This improvement stems from modeling based on human hand priors, bringing the unified dexterous hand closer to human grasping patterns and aligning gestures with functional intentions, further validating the method's effectiveness and stability in functional grasping tasks.}

\begin{figure}[!t]
    \centering
    \includegraphics[width=\linewidth]{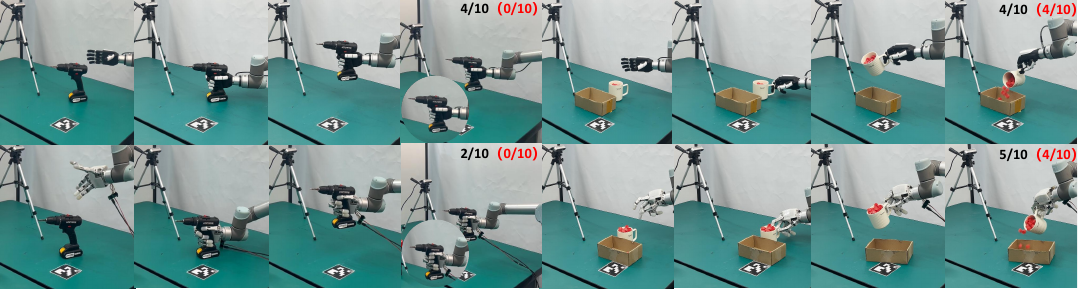}
    \caption{\changed{Real-world functional manipulation results in multiple robotics hands. DFG~\cite{hang2024dexfuncgrasp} results are shown in brackets.}}
    \vspace{-15pt}
    \label{fig:experiment3}
\end{figure}

\textbf{Qualitative Analysis:} 
\changed{Based on the quantitative analysis, we visualized the test gestures under three settings: without condition, One-hot identity encoding, and hand point cloud encoding. Fig.~\ref{fig:Ablation_view} shows that hand point cloud conditioning captures common geometric features across different hands, enabling unified human-like functional gestures. As shown in Fig.~\ref{fig:comparison}, compared to natural human hand motions and DFG~\cite{hang2024dexfuncgrasp}, our method localizes gestures more accurately at functional contact regions while maintaining stable finger envelopment, and executes tasks such as pressing the drill button (6/20) or pouring (9/20) with higher precision and reliability in Fig.~\ref{fig:simulation} and Fig.~\ref{fig:experiment3}.}

\changed{In contrast, DFG~\cite{hang2024dexfuncgrasp} depends purely on network prediction without structural or motion priors, resulting in less reliable finger coordination. Our method achieves more stable and human-consistent dexterous performance.}

\subsection{Real-World Experiments}\label{Real-world Experiments}

In real-world experiments, we adopted a cost-effective setup combining a UR5 robotic arm with the InspireHand and \changed{HnuHand~\cite{10011809}} for functional grasping validation. 
 \begin{wrapfigure}{r}{0.21\textwidth} 
    \centering
    \includegraphics[width=0.21\textwidth]{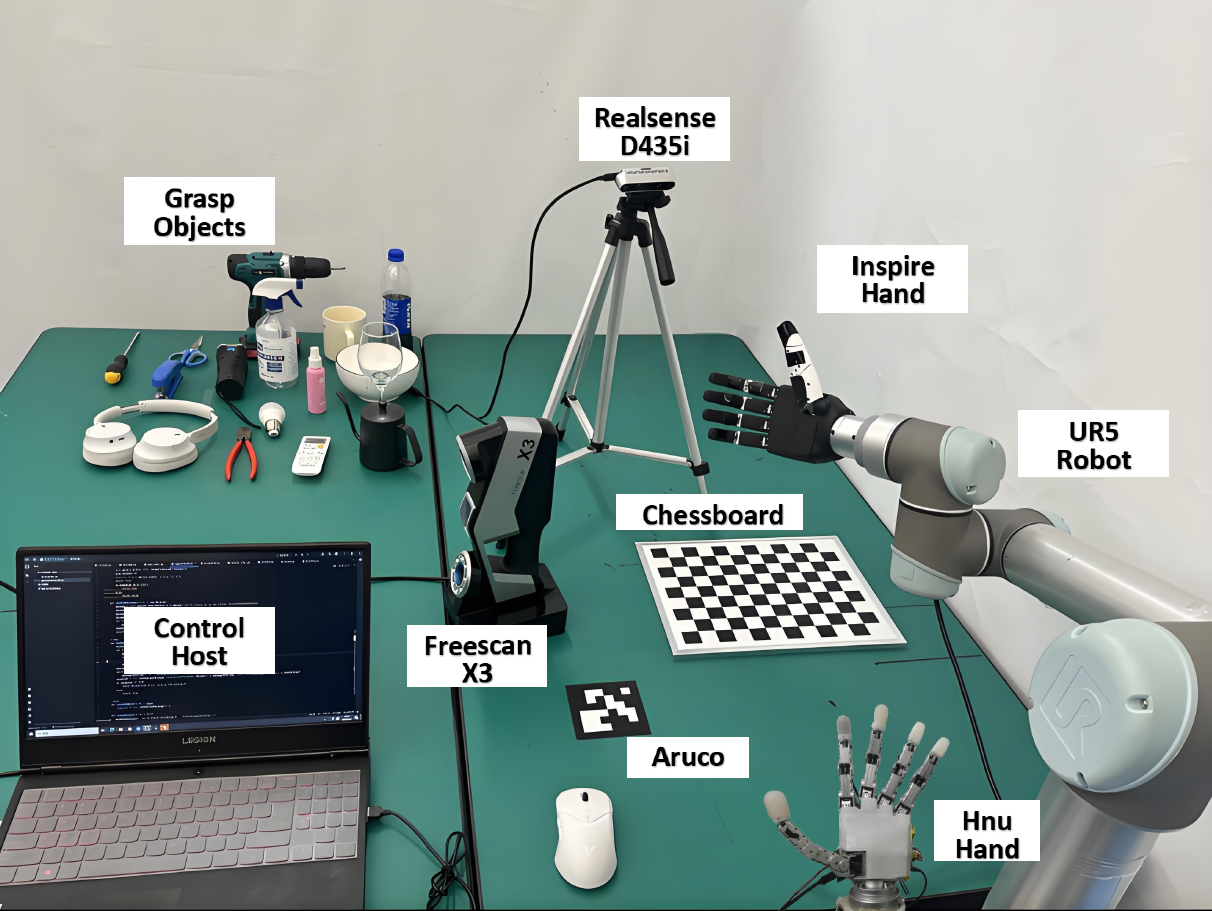}
    \caption{Real-world experiment setting.}
    \vspace{-10pt}
    \label{fig:experiment2}
\end{wrapfigure}As illustrated in Fig.~\ref{fig:experiment2}, the platform consists of \changed{two dexterous hands}, a UR5 arm, a RealSense camera, a calibration board, Aruco codes, a 3D scanner, and a control computer. We first scanned several target objects (\textit{e.g.}, \textit{bottle}, \textit{drill}, \textit{spraybottle}, \textit{flashlight}, and \textit{mug}) using a FreeScan X3 scanner for modeling and post-processing. After calibrating the intrinsics and extrinsics of the RealSense camera, object poses were estimated using FoundationPose~\cite{Wen_Yang_Kautz_Birchfield_2023}.
Uniform point cloud sampling and registration were performed on object surfaces. The processed point clouds were input to the functional grasp model, generating end-effector poses \(R, T\) and gesture parameters \(Q\). Thanks to \changed{the two robotics hands’} coupling, only relevant active joints are controlled, with success criteria consistent with simulation. 
\changed{After the grasps, minor predefined joint and rotation changes were applied to verify gesture stability and their effectiveness for subsequent functional tasks, such as pressing a button or pouring water.} As shown in Table~\ref{tab:real_eval}, \changed{Our method outperformed the latest DFG~\cite{hang2024dexfuncgrasp} across five unseen object categories, improving the success rate by $11.54\%$ on InspireHand and $23.08\%$ on HnuHand~\cite{10011809}.} Moreover, as shown in Fig.~\ref{fig:experiment3}, \changed{our method, evaluated on different dexterous hands after successful grasps, outperformed DFG~\cite{hang2024dexfuncgrasp} by over $75\%$ on key functional tasks (\textit{e.g.}, \textit{press button} and \textit{pour water}), demonstrating more precise control of functional regions via hand motion priors. Despite the complexity of dexterous hands limiting overall success, our results show that reliable, human-like motions can improve functional grasping generalization.}

\begin{table}[!t]
\centering
\footnotesize
\caption{\changed{Real-world functional grasping and manipulation results on different robotic hands.}}
\resizebox{\linewidth}{!}{%
\begin{tabular}{@{}lcccccccc@{}} %
\toprule
\textbf{Method} & \textbf{Bottle} & \textbf{Spraybottle} & \textbf{Flashlight} & \textbf{Drill} & \textbf{Mug} & \textbf{Press Button} & \textbf{Pour Water} & \textbf{Total} \\ 
\midrule
\textbf{DFG~\cite{hang2024dexfuncgrasp}} & 10/10 & 4/10 & 8/10 & 0/10 & 4/10 & 0/10 & 4/10 & 30/70 \\
\textbf{Ours (Inspire)} & 8/10 & 5/10 & 6/10 & 5/10 & 5/10 & 4/10 & 4/10 & 37/70 \\
\textbf{Ours (Hnu)} & 7/10 & 6/10 & 8/10 & 5/10 & 6/10 & 2/10 & 5/10 & 39/70 \\
\bottomrule
\end{tabular}%
}
\vspace{-12pt}
\label{tab:real_eval}
\end{table}

%% file: tex/5-conclusion.tex
This work presents an efficient human hand mapping strategy, modeling hand motions as a sparse matrix to enable human-centered, real-time functional gesture transfer from robots to human hand. Combined with geometric force closure analysis, it effectively evaluates grasp stability. Based on this, we built a large-scale functional grasp dataset. Experiments show our strategy and dataset accurately capture grasp quality, supporting diverse and stable grasps that outperform existing methods. Real-world tests indicate that, while reliable functional grasps can be achieved on unseen instance, hand differences pose challenges for broader generalization. Future work will integrate physics simulation and multimodal sensing to further improve gesture precision, multi-hand generalization, and grasp performance.